\documentclass[sigconf, nonacm, screen]{acmart}

\usepackage[noend]{algpseudocode}
\usepackage{float}
\usepackage{subcaption}
\usepackage{tabularx}
\usepackage{algorithm2e}
\usepackage{multirow}
\usepackage{booktabs}
\usepackage{tablefootnote}
\usepackage{threeparttable}
\usepackage{placeins}
\usepackage{threeparttable}
\usepackage{tcolorbox}
\usepackage{wrapfig}

\usepackage{amsmath,amssymb}
\usepackage{pifont}
\usepackage{xcolor}
\usepackage{listings}
\usepackage{booktabs}
\usepackage{syntax}
\definecolor{codegreen}{rgb}{0,0.6,0}
\definecolor{codegray}{rgb}{0.5,0.5,0.5}

\definecolor{backcolour}{RGB}{245,248,250}
\definecolor{emph}{RGB}{166,88,53}
\definecolor{nightblue}{RGB}{9,49,105}
\definecolor{keywords}{RGB}{207,33,46}
\definecolor{lightpurple}{RGB}{130,81,223}

\lstdefinestyle{mystyle}{
    backgroundcolor=\color{backcolour},   
    commentstyle=\color{codegreen},
    keywordstyle=\color{keywords},
    stringstyle=\color{nightblue},
    basicstyle=\ttfamily\footnotesize,
    breakatwhitespace=false,         
    breaklines=true,                 
    captionpos=b,                    
    keepspaces=true,                 
    numberstyle=\small\color{codegray},
    numbers=left,                    
    numbersep=5pt,   
    xleftmargin=0.2cm,
    aboveskip=0.2cm,
    belowskip=0.1cm,
    showspaces=false,                
    showstringspaces=false,
    showtabs=false,                  
    tabsize=2,
    frame=shadowbox,
    emph={},
    emphstyle={\color{lightpurple}},
}

\definecolor{pastelyellow}{RGB}{255, 255, 230}
\definecolor{lightorange}{RGB}{255, 223, 186}
\definecolor{orange}{RGB}{255, 140, 0}

\AtBeginDocument{%
  \providecommand\BibTeX{{%
    \normalfont B\kern-0.5em{\scshape i\kern-0.25em b}\kern-0.8em\TeX}}}

\begin{document}

\title{Asynchronous Tool Usage for Real-Time Agents } 

\author{
    Antonio A. Ginart*\textsuperscript{} \quad 
    Naveen Kodali*\textsuperscript{} \quad 
    Jason Lee*\textsuperscript{} \quad 
    \\ Caiming Xiong\textsuperscript{} \quad 
    Silvio Savarese\textsuperscript\quad 
    John Emmons*\textsuperscript{} \\
    \normalsize \textsuperscript{}Salesforce AI Research \quad
}

\thanks{*Core contributors}

\begin{abstract}
\small
While frontier large language models (LLMs) are capable tool-using agents, current AI systems still operate in a strict turn-based fashion, oblivious to passage of time.
This synchronous design forces user queries and tool-use to occur sequentially, preventing the systems from multitasking and reducing interactivity.
To address this limitation, we introduce \textit{asynchronous} AI agents capable of parallel processing and real-time tool-use.
Our key contribution is an event-driven finite-state machine architecture for agent execution and prompting, integrated with automatic speech recognition and text-to-speech.
Additionally, we present the first dataset tailored for instruction-tuning LLMs for asynchronous tool-use. Drawing inspiration from the concepts originally developed for real-time operating systems, this work presents both a conceptual framework and practical tools for creating AI agents capable of fluid, multitasking interactions.
\normalsize
\end{abstract}

\maketitle

\section{Introduction}

With the advancement of large-scale foundation models \cite{bommasani2022opportunitiesrisksfoundationmodels} into the terabyte realm, AI models have become sufficiently capable to function as tool-using agents \cite{dubey2024llama3herdmodels}.
A crucial component in training these base models to  user instructions is supervised instruction tuning \cite{dubey2024llama3herdmodels, ouyang2022traininglanguagemodelsfollow, peng2023instructiontuninggpt4}.
In addition, the integration of tool-use datasets serves to further enhance the capabilities of these agents \cite{zhang2024xlamfamilylargeaction}. 
Despite these advancements, there is, to our knowledge, an underlying assumption that persists in the design of leading AI agents and the instruction datasets they are trained on: these agents are inherently \textit{synchronous}.
This synchronous nature becomes evident through user interactions, such as watching a loading spinner when an agent employs a tool.
Giving agents the ability to (1) manage multiple concurrent processes in real-time, and (2) asynchronously respond to the user as soon as any process finishes will significantly reduce perceived delay and meaningfully improve user experience.

Consider the following hypothetical scenario of a voice call with an AI travel concierge (see Fig. 1).
At a glance, this interaction may seem uninteresting.
However, consider that preparing a detailed travel itinerary might be a multi-step and multi-tool task that may take, say, 30 seconds to fully prepare.
Thus, it is preferable for the concierge to quickly verbally reply rather than wait to respond until the itinerary is ready.
Furthermore, the user asked for a second tool-use request while still waiting on the first.
While humans have the ability to maintain multiple concurrent thought processes simultaneously, current frontier agents are fully serial\footnote{See, for example: \texttt{chatgpt.com}, \texttt{claude.ai}, or \texttt{perplexity.ai}}.
When a serial agent is in the middle of processing a user request that involves long-running tools or reasoning chains, the user has no choice but to abort the request or wait for it to finish before being able to submit a subsequent query.

\begin{figure}
\begin{tcolorbox}[colframe=black, left=2pt, right=2pt, top=2pt, bottom=2pt, title=, label={box:dialogue}]
\small
\textsc{User:} \textit{Please present detailed travel itinerary for my trip to Miami next week.}

\textsc{Concierge:} \textit{Certainly! I will prepare this for you momentarily.} 

\textsc{User:} \textit{Also, what's the weather going to be like?}

\textsc{Concierge:} \textit{Based on current forecasts, the weather in Miami next week is expected to be warm and humid, with daytime highs around 88°F and overnight lows around 76°F. There's a chance of afternoon thunderstorms on several days, which is typical for Miami this time of year. I'd recommend packing light, breathable clothing, sunscreen, and a light rain jacket or umbrella. Would you like me to include some indoor activity options in your itinerary in case of rain?}

\textsc{User:} \textit{Sounds great!}

\textsc{Concierge:} \textit{Here is your itinerary, including indoor activity options...}
\end{tcolorbox}
\normalsize

    \caption{Hypothetical voice call between a user and a travel agent with \emph{asynchronous} tool use.}

\end{figure}

While frontier models are indeed highly general, and are in fact able to solve a variety of tasks zero-shot, including various forms of tool usage, we find that even with specialized prompting, frontier models struggle to operate in an asynchronous fashion under certain circumstances. For example, while specialized prompting does enable frontier LLMs to operate asynchronously to some degree, they may still get confused in certain scenarios with out-of-order messages. 

Quite recently, OpenAI has released a real-time voice API\footnote{\texttt{openai.com/index/introducing-the-realtime-api/}} that supports asynchronous tool usage. However, few technical, architecture, or model details have been shared. To our knowledge, there is no other system supporting real-time voice and asynchronous tool usage.

Other available frontier LLMs  seem to lack a precise notion of time and a perfect understanding of asynchronous messages. It is hard to address this problem today due to a lack of instruction tuning data for asynchronous tool usage. Furthermore, the machine learning systems co-design aspects of this problem requires some specific familiarity across both subfields.

In this work, we draw inspiration from the literature of concurrent programming and asynchronous computer system in order to present a conceptual framework and practical architecture for building asynchronous agents with frontier language models.


We propose and implement a novel event-driven finite-state machine architecture for executing and prompting AI agents in asynchronous tool environments. The system, herein called the execution environment, is complete with peripherals such as automatic speech recognition and text-to-speech. Our system is modular, and can work with any LLM that generates valid messages. We fine-tune both Llama 3.1 and GPT-4o to operate the execution environment and find compelling performance from both models. We discuss and characterize trade-offs in event-driven AI architectures, such context management via forking vs. spawning.


    
  


\section{Background \& Related Works}

We provide some background, with a focus on more fundamental and seminal works on the systems side (given that we adopt classically established systems paradigms) and more contemporaneous, bleeding-edge works on the generative machine learning side (given that this field is dynamic and advancing rapidly).

\subsection{Asynchronous Computer Systems}

\subsubsection{Asynchronous Execution}
Asynchronous execution, a cornerstone of our proposed framework for AI agents in tool environments, has its roots in the seminal work of Dijkstra (1965) on cooperating sequential processes \cite{dijkstra2002cooperating}.
Hoare \cite{hoare1978communicating} offered a formal framework for describing and analyzing asynchronous systems, which informs our approach to designing AI agent interactions in asynchronous environments.

\subsubsection{Polling v. Interrupt-Based Concurrency}
In developing our conceptual framework for AI agents in asynchronous tool environments, we draw upon the long-standing debate between polling and interrupt-based approaches to concurrency. \cite{hansen1973operating} offers insight into this trade-off in the context of operating systems. Lampson and Redell's \cite{lampson1980experience} work on processes and monitors in Mesa offers insights into the trade-offs impacting a system's responsiveness to asynchronous events, such as automatic speech recognition inputs. However, perhaps the most influential and relevant prior work is the Robot Operating System (ROS) \cite{koubaa2017robot}, which opts for an event-driven concurrency model.

\subsubsection{Real-Time Systems}

Real-time computing systems, and operating systems in particular, is an established branch of computer engineering that encompasses systems building for real-time timing requirements in a concurrent environment with integrated peripheral sensors and actuators. We include references to recent surveys \cite{rtos, stankovic2004real, meghanathan2005survey, darby2011rto} and classical texts \cite{rogers2001real, rammig2009basic, sched, stankovic1989spring}.

\subsection{Generative AI}


\subsubsection{Large Action Models and Tool-Use}

Over the past three years, as LLM model weights and training data grew from the gigabyte to terabyte scale, LLM use-cases expanded from simple chatbots to helpful copilots all the way to now, tool-using autonomous agents. 

A recent study introduced xLAM, a series of large action models specifically designed for AI agent tasks, ranging from 1B to 8x22B parameters \cite{zhang2024xlamfamilylargeaction}. The xLAM models demonstrate exceptional performance across multiple agent ability benchmarks, notably outperforming GPT-4 and Claude-3 in function calling tasks, and aim to advance open-source LLMs for autonomous AI agents. APIGen, an automated pipeline for generating high-quality function-calling datasets, leverages over 3,600 APIs across 21 categories for data generation \cite{liu2024apigenautomatedpipelinegenerating}.

\subsubsection{Multi-Agent Systems}

Multi-agent systems in the context of generative AI have seen significant advancements in recent years.rary challenges and opportunities in developing AI systems that can effectively collaborate. The literature on cooperating multi-agent AI systems is vast, but some particularly important or relevant works from the before the foundation model era include \cite{torreno2017cooperative,  yu2022surprising, kraus1997negotiation, jennings1995controlling, dorri2018multi}.

More recently, the advent of generative foundation models has ushered in a new era of AI coordination amongst LLM-based agents. Development of platforms like AutoGen \cite{wu2023autogen} provides practical tools for implementing and experimenting with multi-agent AI systems, facilitating research and applications in this rapidly evolving field.

Diversity Empowered Intelligence (DEI) is framework that leverages the diverse expertise of multiple software engineering agents to enhance problem-solving capabilities.  Experimental results demonstrated that a DEI-guided committee of agents significantly outperformed individual agents \cite{zhang2024diversityempowersintelligenceintegrating}.

AgentLite is lightweight open-source library designed to simplify the development and evaluation of LLM-based agent systems \cite{liu2024agentlitelightweightlibrarybuilding}. AgentLite offers a task-oriented framework that enhances agents' ability to break down tasks and facilitates multi-agent system development, providing researchers with a user-friendly platform for innovating LLM agent reasoning strategies and architectures.

Unlike prior work, our work is far more focused on the asynchronous and real-time aspects of LLM-based agents rather than the cooperation and coordination aspects of multi-agent systems. Many of the ideas in the prior multi-agent systems literature are in fact quite complimentary and could be translated to the asynchronous setting. Other relevant works on multi-agent generative systems include \cite{guo2024largelanguagemodelbased, liu2023dynamicllmagentnetworkllmagent, guo2024embodiedllmagentslearn, talebirad2023multiagentcollaborationharnessingpower, dong2024multi, zhang2024efficientllmgroundingembodied}.

\subsubsection{Speech Models}

Speech models have undergone remarkable advancements in recent years. The introduction of Whisper \cite{radford2023robust} marked a significant milestone in automatic speech recognition (ASR), demonstrating robust performance across multiple languages and accents.

In the domain of text-to-speech (TTS), models like VALL-E 2 \cite{chen2024vall} have shown impressive capabilities in voice cloning and style transfer, requiring only short audio samples. Voicebox \cite{le2024voicebox} has demonstrated state-of-the-art performance in various speech generation tasks, including noise removal, content editing, and cross-lingual style transfer. NaturalSpeech \cite{tan2024naturalspeech} approaches human-level quality in speech synthesis.

These advancements are paving the way for more natural and versatile speech interfaces in AI systems, with potential applications ranging from virtual assistants to accessibility tools.

\subsubsection{Spoken Dialogue Systems}

Spoken dialogue systems have seen significant advancements in recent years, particularly in the areas of dialogue state tracking and dialogue management, due to rapid development in natural language processing powered by LLMs. \cite{wahlster2023understanding, gunasekara2024overview, jannach2023evaluating} provide details on recent developments in this subfield.

\section{Real-Time Agents}

The proposed framework for real-time agents combines an asynchronous execution environment with a markup language prompting specification. Similar to a software-hardware divide, as long as the LLM produces valid generations according to the specification, the environment will asynchronously enqueue function calls\footnote{We use the terms functions and tools interchangeably.} and handle chat interactions through speech-to-text and text-to-speech peripherals.

The asynchronous execution environment is, at its core, an event-driven finite state machine \cite{harel1987statecharts}, augmented with priority scheduling (via priority queue) \cite{miller1960priority}. We refer to this core component of the execution environment as the \textit{dialog system}, comprised of the \textit{dialog FSM} and the \textit{scheduling queue}. The LLM generation, context management, and function calling are handled by the \textit{dispatcher}, which is comprised of the \textit{dispatch language model} and the \textit{ledger}, which acts as the single-source-of-truth for the dispatch LM's context. Messages are appended to the ledger atomically, generally as a consequence of the dialog FSM processing the schedueling queue, but there are exceptions, such as for handling user interruptions. Refer to Fig. 2 for a schematic of the execution environment with speech-to-text (STT) and text-to-speech (TTS) I/O. While real-time voice agents are an important motivation, they are not the only use case for real-time asynchronous tool usage, so the STT and TTS messages could be replaced with input and output text streams.

\begin{figure}
    \centering
    \includegraphics[width=0.47\textwidth]{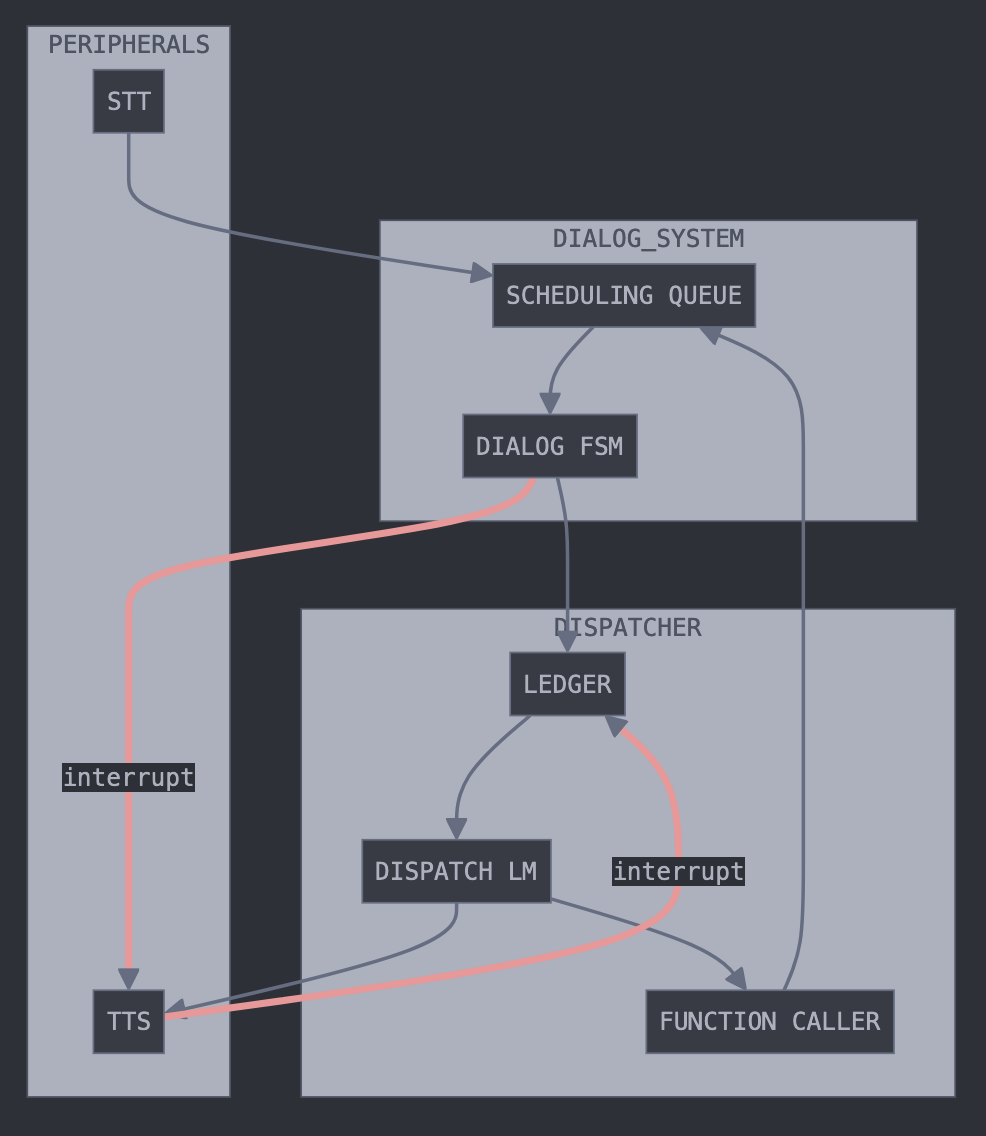}
    \caption{Architecture diagram for the asynchronous execution environment with voice peripherals}
    \label{fig:architecture}
\end{figure}

\subsection{Architecture}

\subsubsection{Clock Awareness}

One important aspect of our paradigm arises from the usage of timestamps in messages, clock update messages at discrete time intervals (for example, every 5 seconds) coupled with a notion of clock awareness present in the LLM (due to specialized fine-tuning or prompting).

\subsubsection{Parallel Thought Processes: Fork vs. Spawn Semantics}

We propose a technical definition of a \textit{parallel thought process}. Analagous to a parallel programmatic process in a traditional software sense, parallel thought process is an concurrent instance of the asynchronous execution environment with parent-child semantics. The child's input stream is populated by function calls from the parent and the child's output stream populates the function responses to the parent. Concretely, parallel thought processes are created via either \texttt{fork} or \texttt{spawn} calls. For a \texttt{fork} call, the parent initializes the child's ledger with a copy of its own and appends a new message containing further instructions for the child. For a \texttt{spawn} call, the parent initializes the child with a new ledger and populates the first message containing the child's instructions. It is up to the parent thought processes to determine if a \texttt{fork} or a \texttt{spawn} call is more appropriate on a case-by-case basis, given that there are clear trade-offs for each type of parallel thought process. Forking uses more context in the child and therefore could be more expensive while also including potentially unnecessary or distracting messages, so it should only be used if the child needs a full view into the parent's context in order to achieve its goal. By default, spawning is probably preferable for most cases, since the parent can usually summarize the relevant details into the child's instructions. By default, the dispatch LM could be the same as the parent, but in principle, nothing prevents the parent from prescribing a different dispatch LM (as long as it's fine-tuned or prompted to correctly handle prompt template as expected by the environment). For both \texttt{fork} and \texttt{spawn}, we have a third reserved tool, \texttt{kill}, that the dispatching LM can use to interrupt and terminate parallel thought process. As is implied by these function call semantics, there is the possibility of recursive creation of parallel thought processes, which could be used to dynamically organize multi-agent hierarchies at runtime. Of course, custom-built tool calls can also serve this purpose.

\subsubsection{Event Processing} The dialog system implements an event-driven FSM with priority scheduling. An event will contain a priority level and (potentially) cause a state transition to occur. Some events also contain messages to be appended to the ledger. Events can be produced by:

\begin{enumerate}
    \item The STT (or input peripheral), when the user begins speaking or finishes speaking
    \item The dispatcher, when the dispatch LM begins or finishes generation
   \item  The TTS (or output peripheral), when the output stream begins and finishes emitting
   \item The function caller, when a tool-use request is sent or a response is received
\end{enumerate}

In order to ensure that the dialog FSM state variable accurately reflects the overall system, state transitions push from the dispatcher and the peripherals should essentially always have the minimum possible priority, $-\infty$, in order to ensure they are processed instantly. Alternatively, the dispatcher and the peripherals can use a locking mechanism to atomically update the state variable when appropriate (essentially skipping the scheduling queue altogether). Function call responses, on the other hand, should never use such a priority, and instead should use a developer defined priority, always going through the scheduling queue.    push events to the scheduling queue based on on internal processing or state change. As a concrete example, the execution environment must ensure that the TTS is halted if the interrupt event is pushed by the STT subsystem because the user starts to speak.

\normalsize
\vspace{5pt}

\small
\begin{tabular}{|c|c|c|c|}
    \hline
    \textbf{Event} & \textbf{Priority} & \textbf{Message} & \textbf{State} \\
    \hline
    generate_done & $-\infty$  & No  & Idle \\
    \hline
    emit & $-\infty$ & No & Emitting \\
    \hline
    emit_done & $-\infty$ & No & Idle \\
    \hline
    interrupt & $-\infty$ & Yes & Listening \\
    \hline
    tool_response_received & $p$ & Yes & Generating \\
    \hline
    user_chat & $-1$ & Yes & Generating \\
    \hline
    tool_request_sent & $-\infty$ & Yes & Idle \\
    \hline
    time_passage & $1$ & Yes & Generating \\
    \hline
\end{tabular}
\vspace{2pt}

\textbf{Table 1:} \textbf{ Events. } The Priority column is the priority level in the scheduling queue ($p$ denotes a tool-specific variable-defined priority level). The Message column is \textit{Yes} if the event contains a message to be appended to Ledger and \textit{No} otherwise. The State column contains the next state for the FSM to transition into. The FSM has four states: \textit{idle}, \textit{listening}, \textit{generating}, and \textit{emitting}. There exist certain environment conditions under which the emit done event would actually transition to a generating state. See Section 3.3.3 for more details.

\normalsize

\subsection{Peripherals}

We use the open-source Pipecat\footnote{\texttt{github.com/pipecat-ai/pipecat}} framework to integrate our LLM with STT and TTS peripherals. This framework enables real-time voice recognition and synthesis, and helps handle user interruptions. It does not support asynchronous tools, however, which we implement in our system. LLM generations are streamed into a sentence aggregator and then processed phrase-by-phrase by the TTS. We observe end-to-end latency of <$300$ms, which is enough to mimic human conversation.

\subsubsection{Speech-to-Text}

There are various suitable choices for STT services, both among the open-source packages and APIs. We found Deepgram\footnote{\texttt{deepgram.com}} to be a good API option. We also obtained low-latency STT with an optimized implementation of Whisper Turbo\footnote{\texttt{huggingface.co/openai/whisper-large-v3-turbo}}.

\subsubsection{Text-to-Speech}

We found that TTS had a slightly higher variation in latency and quality among leading open-source packages and APIs. We preferred the Sonic API\footnote{\texttt{cartesia.ai/sonic}} for its speed and quality.

\subsection{Execution} 
OpenAI introduced \textit{chat markup language} (ChatML)\footnote{\texttt{github.com/openai/openai-python/blob/release-v0.28.0/chatml.md}} as the original prompting template for fine-tuned chat models. ChatML originated as a specific markup language that implement an abstract data type (ADT), but, interestingly, it was the ADT implied by ChatML that has actually caught on as the universal standard interface to chat models for developers across all major APIs. Today, the ChatML interface enables developers to more easily manipulate LLM context without resorting to boilerplate string manipulations.

We propose an extension to ChatML for asynchronous real-time agents. However, beyond just being an interface, we actually view this as closer to an \textit{instruction set} because it provides an abstraction layer between the chat model and the asynchronous execution environment. Any chat model that generates valid (asynchronous) ChatML, should, in principle, be able to run on any asynchronous execution environment that implements the specifications.

\subsubsection{Prompt Template and Context Management}

We review the standard practice for synchronous chat model context, here denoted by $\mathbf{C}$. Precisely, context $\mathbf{C}$ is a list of length $\ell$ of message dictionaries, $m_i$, with each message dictionary containing role ($\texttt{role}_i$) and content ($\texttt{content}_i$) fields.

$$ \mathbf{C}= \left((\texttt{role}_1, \texttt{content}_1), ..., (\texttt{role}_\ell, \texttt{content}_\ell)\right)$$

where $\texttt{role}_i \in \{ \texttt{system}, \texttt{assistant}, \texttt{user}\}$ and  $\texttt{content}_i \in \Sigma^*$ where $\Sigma$ denotes the token vocabulary for the language model (including the empty string). We do not impose any restrictions about the number or order of the roles in the context. For example, we allow back-to-back \texttt{user} messages.

In the asynchronous case, the ledger, $\mathbf{L}$ is similarly a list of messages, $ (\texttt{role}_i, \texttt{content}_i)$. We include a set of tools, denoted $\mathcal{T}$, modeled via $\textit{tool-use function}$ $f: \texttt{JSON} \rightarrow \Sigma^*$, which inputs valid JSON, selects and executes the corresponding tool in $\mathcal{T}$ with the corresponding arguments, and outputs a response string. We also include certain \textit{reserved tools}, including \texttt{fork}, and \texttt{spawn} and \texttt{kill}. We include a new, fourth role, $\texttt{notification}$ that is managed by the the asynchronous execution environment. It will also be useful to define a set of hash strings $\mathbb{H}$.

If $\texttt{role}_i = \texttt{user}$
$$ \texttt{content}_i = (\texttt{timestamp}_i, \texttt{chat}_i)$$

where $\texttt{timetamp}_i \in \mathrm{N}$ (or some other suitable timestamp format) and $\texttt{chat}_i \in \Sigma^*$.

If $\texttt{role}_i = \texttt{assistant}$
$$ \texttt{content}_i = (\texttt{thought}_i, \texttt{function}_i, \texttt{chat}_i)$$
where $\texttt{thought}_i \in \Sigma^*$ and $\texttt{function}_i \in \mathbf{List}[\texttt{JSON}]$ is a list of valid JSONs.

If $\texttt{role}_i = \texttt{notification}$
$$ \texttt{content}_i = (\texttt{source}_i, \texttt{timestamp}_i, \texttt{data}_i)$$

The $\texttt{source}$ field 
where $\texttt{source}_i \in \mathcal{T} \times \mathbb{H}$ and $\texttt{thought}_i \in \Sigma^*$. We do not change the content or format for \texttt{system} messages.



Like with synchronous chat markup language, the choice of special tokens and details of the prompting template to serialize and deserialize the abstraction is specific to the language model's vocabulary.

Frontier models now offer context lengths upwards of 128k tokens, which can enable voice conversations with tool calls lasting up to 30 minutes or more. Context length and memory management beyond this is an interesting topic which we do not delve into here.

\subsubsection{Instruction Set} We describe how the asynchronous execution environment should handle messages. Like an instruction-tuned chat model, it is the dispatch LM's burden to ensure it generates valid messages with respect to the instruction set.

The dispatch LM is the only component of the system that generates \texttt{assistant} messages. For the \texttt{thought} field the environment does nothing. This is just an LLM scratchpad. For the \texttt{chat} field, the environment is responsible for initializing a real-time output stream to the user (either via a text-to-speech peripheral or a token stream directly). Finally, the environment should parse the \texttt{function} field and asynchronously execute the requests. Note that, in our system, token generation and TTS emitting happen concurrently, and it is technically possible for TTS emitting to finish before the generation of the assistant message. For example, consider the case in which an assistant message begins with a \texttt{chat} and then includes a \texttt{thought}. While the message as an abstract date type is a dictionary, when implemented, it streams in serially and we process it in real-time as a stream. If a chat streams in, the TTS component will start emitting sentence-by-sentence, meaning the TTS will run in a delay, but it is possible that the TTS could finish emitting the chat before the LLM generation finishes the subsequent thought. We omit the details of this edge case in Table 1 and Section 3.1, but it is important to handle this case because the dialog FSM state should revert back to generating rather than transition to idle in such a situation.

\begin{tcolorbox}[colframe=black, left=2pt, right=2pt, top=2pt, bottom=2pt, title=, label={box:dialogue}]
\small
$$
\begin{aligned}
& \textbf{Algorithm 1:} \textbf{ Handling Events } \\
& \textbf{Input:} \text{ Queue } Q, \text{ Environment State } E \\
& \textbf{Output:} \text{ Updated Ledger } L, \text{ Updated State } E \\
& p \leftarrow Q.\text{top}.\text{priority} \\
& run \gets E = \text{idle} \textbf{ or } \\ & \quad ( E = \text{generating} \textbf{ and } \text{ $p \leq 1$} ) \textbf{ or } \\ & \quad (E = \text{emitting} \textbf{ and } \text{ $p < 1$}) \\
& \textbf{if } run \textbf{ then} \\
& \quad \text{event} \leftarrow Q.\text{pop}()   \\
& \quad L.\text{append}(\text{event.message}) \text{ // No-op if message is empty}\\
& \quad E \leftarrow \text{event.state} \\
& \textbf{else if } E = \text{listening} \textbf{ then} \\
& \quad \textbf{pass } \text{ // optionally, we could interrupt the user if $p < -1$ }\\
& \textbf{return } L, E
\end{aligned}
$$

\normalsize
\end{tcolorbox}

Upon the successful start of the function call (for example, asynchronously sending an HTTPS request), the execution environment  should append a \textit{request-sent} notification message. This message should include the name of the tool used, a request ID, as well as potentially other metadata such as the arguments. For example:

\vspace{5pt} 

$\textbf{Message: } j$

$\texttt{role} = \texttt{notification}$

$\texttt{source} = \texttt{system}$ 

$\texttt{data} = \texttt{"Request sent for: search. ID: 0abd754d495."}$ 
\vspace{5pt} 

Upon the completion of the $j$-th function call from the $i$-th message, the execution environment should append a \textit{response-received} notification message with $\texttt{data}  \gets f(\texttt{function}_{ij})$. For example:

\vspace{5pt} 

$\textbf{Message: } j+1$

$\texttt{role} = \texttt{notification}$

$\texttt{source} = (\texttt{search}, \texttt{"0abd754d495"})$ 

$\texttt{data} = \texttt{"Here are your results..."}$ 
\vspace{5pt}

The \texttt{user} messages are essentially unchanged and still represent direct input from the user. The environment adds timestamps. 

\vspace{-5pt}

\subsubsection{Interruption Handling} 

At any given moment in time, the FSM is in one of four states with respect to the execution environment: \textit{idle}, \textit{generating}, \textit{emitting}, or \textit{listening}. The implementation of the execution environment as a whole is responsible for ensure the FSM state is accurately reflected in the true state of the system.

For example, if the TTS is streaming output (either voice or text) to the user, then the FSM is emitting. Otherwise, if it is generating tokens but not emitting, then it is generating. If the user is in the process of creating input (for exampling, speaking), then the FSM is listening. Else, the FSM is said to be idle.

Interruptions are a first-class feature in an asynchronous agent. We explicitly include interruptions as part of the proposed instruction set. The scheduling queue allows for the environment to enforce atomic updates to the ledger despite the concurrency. Every 5 seconds, the system queues up a time passage notification message.

All messages have a priority\footnote{Inherited from the priority of the event containing said message}. As example defaults, \texttt{user} messages are priority \texttt{-1} and \texttt{assistant} messages are priority \texttt{1} (although this can be configured on a per-deployment basis).

The key difference in how interruptions are handled in the \textit{generating} and \textit{emitting} states is that for a tie in priority level, the interrupt occurs if the dispatch LM generating but does not if it is emitting.

Tool definitions should include a priority, but the default priority for request-sent and response-received messages is 1. 

The execution environment is responsible for correctly bookkeeping the ledger during interruptions. Consider the following example.

\vspace{5pt} 

$\textbf{Message: } l$

$\texttt{role} = \texttt{assistant}$

$\texttt{chat} = \texttt{"Blah blah blah <|interrupt|>"}$ 

$\textbf{Message: } l+1$

$\texttt{role} = \texttt{notification}$

$\texttt{source} = \texttt{system}$ 

$\texttt{data} = \texttt{"Assistant interrupted due to user speaking"}$

\footnote{Note that message $l+1$ is posted to the ledger immediately when the user starts speaking, whereas message $l+2$ is posted once the user is done.}$\textbf{Message: } l+2$

$\texttt{role} = \texttt{user}$

$\texttt{chat} = \texttt{"I am interrupting you."}$ 

\vspace{5pt} 
We add a special interruption token, as depicted above. Note that, due to the pipeline nature of the asychronous execution environment, there is a time delay between token generation and the output stream (for example, via a TTS module). The system is responsible for making sure the number of \textit{blah} accurately reflects how many the assistant said before interruption. In this case, the LLM may have generated a token stream of \texttt{"Blah blah blah blah"} while the TTS has only managed to emit  \texttt{"Blah blah blah"}. In this case, the environment is responsible for reconciliation between these two streams and should only update the ledger to reflect the actual output emitted to the user. This issue is only present in the chat component of the LLM generation, since thought streams do not trigger downstream post-processing and therefore can be interrupted naively. For function call streams, the environment could choose to either atomically include a function call only if it was generated to completion, or simply interrupt a partial function call string. In either case, the dispatch LM should rely on a request-sent message to indicate the completion of the function call request. Below, Fig. 3 offers another perspective on the system implementation, emphasizing the ledger and messages, the control and information flow through various system components and the voice activity detection (we implemented a simple volume based cutoff, see \cite{sohn1999statistical, moattar2009simple, chang2006voice, zhang2012deep, hughes2013recurrent} for more details and more sophisticated approaches).

\begin{figure}
    \centering
    \includegraphics[width=0.47\textwidth]{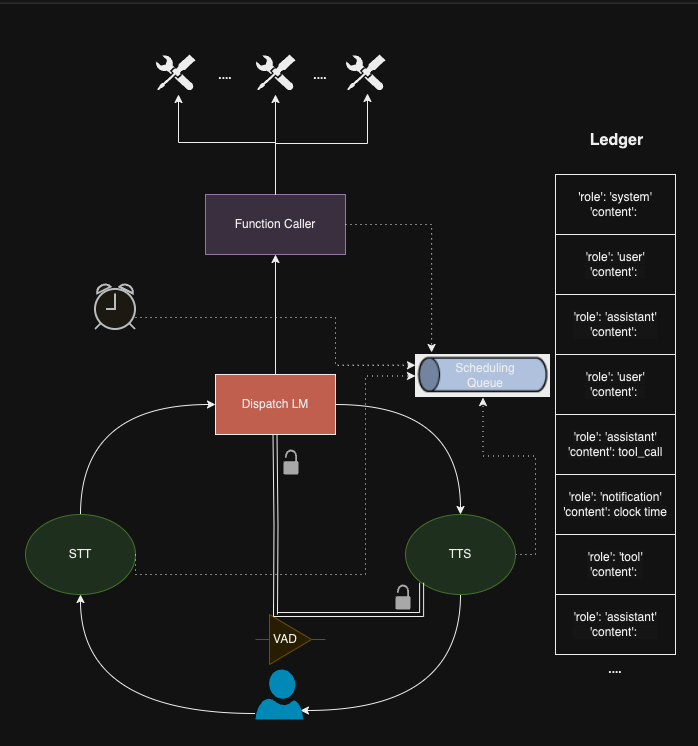}
    \caption{Implementation and control flow diagram for the asynchronous execution environment}
    \label{fig:architecture}
\end{figure}

\section{Discussion}
\paragraph{Results} In the supplementary materials, we share some sample recordings of voice calls with both the fine-tuned Llama 3.1 and GPT-4o as dispatch LMs. Qualitatively, we fine that fine-tuning helps ensure the dispatch LM consistently generates valid ledger messages according to the prescribed instruction set.

\paragraph{Pipelined vs. Multi-modal Systems}

Multi-modal foundation models are an active and promising area of research and development, and are highly relevant to real-time agents, particularly for voice I/O. A foundation model with both text and speech modalities could exhibit tighter integration with peripherals and possibly even further reduce latency or improve speech quality.

In our implementation, we opted for a pipelined system, in which we had independent neural networks for the STT, Dispatch LM, and TTS, which text as the I/O medium between the subnets. However, one could have also used a multi-modal foundation model (as demonstrated by OpenAI). Our architecture is well-suited for such a foundation model, with the exception that the STT and TTS components would not include a neural component, but rather, would simply include a speech tokenizer and vocoder component, respectively. In this case, speech tokens would be the I/O medium between these components.

\paragraph{Reasoning for Time-Constrained Tasks}

One intriguing possibility that is enabled after endowing the agent with a sense of clock awareness is the ability for the user to request a \textit{time-constrained} task. For example: \textit{Complete a research report on state-of-the-art text-to-speech models within the next 10 hours.} Within the proposed framework, the agent's sense of clock awareness enables it to coordinate long-running chain-of-thought (or other multi-step reasoning methods) with multiple iterations of tool-use across, potentially, multiple other task-specific agents while keeping the entire process on a reasonable schedule based on the allotted time limit.

Exploring specific prompt templates and fine-tuning strategies for powering this kind of clock-aware reasoning with time constraints is a fertile topic for future work.

\paragraph{Conclusion and Future Work}

This work introduces an architecture for real-time, asynchronous AI agents capable of fluid, real-time interactions. By implementing an event-driven finite-state machine with asynchronous tool usage and parallel thought processes, we enable more natural and responsive AI interactions, particularly in voice-based applications.

Broadly speaking, we envision a future in which agent LMs are fine-tuned to for precise instruction set specifications which can be executed in any environment supporting said instruction set. Future work should explore integration with multi-modal LMs and seamless integration with multi-agent systems and long-running task-specific agents.

\bibliographystyle{alpha}
\bibliography{sample}

\end{document}